\documentclass{article}
\usepackage{amsmath}

\usepackage{PRIMEarxiv}
\usepackage[utf8]{inputenc} 
\usepackage[T1]{fontenc}    
\usepackage{hyperref}       
\usepackage{url}            
\usepackage{booktabs}       
\usepackage{amsfonts}       
\usepackage{nicefrac}       
\usepackage{microtype}      
\usepackage{lipsum}
\usepackage{fancyhdr}       
\usepackage{graphicx}       
\graphicspath{{media/}}     
\usepackage{subfigure}
\title{GAgent: An Adaptive Rigid-Soft Gripping Agent with Vision Language Models for Complex Lighting Environments}

\author{
\textbf{Zhuowei Li} $^{2,3,4}$ \quad
\textbf{Miao Zhang} $^{1}$$^*$  \quad  \textbf{Xiaotian Lin} $^{2}$   \quad  \textbf{Meng Yin} $^{5}$ \quad \\ 
\textbf{Shuai Lu}  $^{1}$ \quad 
\textbf{Xueqian Wang}  $^{1}$ \\
$^{1}$ Shenzhen International Graduate School,
 Tsinghua University,
$^{2}$ Yongjiang Laboratory,  \\
$^{3}$ University of Nottingham Ningbo China,
$^{4}$ The University of Hong Kong,\\ 
$^{5}$ Shenzhen Institute of Advanced Technology, Chinese Academy of Sciences,\\ 
$^{\ast}$ Corresponding author: Miao Zhang, zhangmiao@sz.tsinghua.edu.cn}

\begin{document}
\maketitle

\begin{abstract}
This paper introduces GAgent: an Gripping Agent designed for open-world environments that provides advanced cognitive abilities via VLM agents and flexible grasping abilities with variable stiffness soft grippers. GAgent comprises three primary components - Prompt Engineer module, Visual-Language Model (VLM) core and Workflow module. These three modules enhance gripper success rates by recognizing objects and materials and accurately estimating grasp area even under challenging lighting conditions. As part of creativity, researchers also created a bionic hybrid soft gripper with variable stiffness capable of gripping heavy loads while still gently engaging objects. This intelligent agent, featuring VLM-based cognitive processing with bionic design, shows promise as it could potentially benefit UAVs in various scenarios.

\end{abstract}


\section{Introduction}
In recent years, the gripping use of unmanned aerial vehicles (UAVs) has emerged as a new trending research direction \cite{lieret2020lightweight,setty2020generic}.
However, the grabbing scenes in the open world are very complex, which leads to the development of robotic grasping systems with advanced cognitive and adaptable grasping capabilities.
To achieve high-level cognitive abilities, reinforcement learning embodiment is studied\cite{kalashnikov2018scalable, nguyen2019review}. In \cite{kalashnikov2018scalable}, Scalable Deep Reinforcement Learning is used to handle large amounts of off-policy image data for complex tasks like grasping. However, RL-based embodiment has posed challenges in terms of generalization capability, sample-effectiveness capability, and profound reasoning capability, especially in dynamic and uncertain real environments.

Recently, Large multimodal models (LMMs), such as MiniGPT-4 \cite{MiniGPT-4} and LLaVA \cite{LLaVA}, have exhibited impressive performance in the domains of natural instruction-following and visual cognition. Therefore, LMMs are integrated with the physical world in the embodied agent. Apart from RL algorithms for specific tasks, LMMs-based agents have generalization capabilities \cite{PaLM-E,liu2023visual} though fine-tune methods, such as human demonstrations \cite{Human_Demonstrations}, vision-language cross-modal connector\cite{Connector}, ever-growing skill library \cite{VOYAGER} and so on. On-policy  (RL) algorithms face challenges in terms of sample efficiency. However, LMMs-based agents, such as \cite{PaLM-E}, benefit from diverse joint training across language, vision, and visual-language domains, demonstrating positive transfer across different tasks and datasets. The ability to reason is also crucial for solving complex tasks. Therefore, LLMs with emerging reasoning abilities \cite{emergent_ability} can be applied to complex tasks in few-shot cases \cite{CoT}.

Although the LLMs-based agent demonstrates many promising potentials, it still encounters environmental challenges in practical applications, especially pixel-level noise such as in low-light scenarios. To the best of the author's knowledge, this is the first attempt to apply an LLMs-based agent to a gripper in a complex light environment.

To address the aspect of adaptable grasping capabilities, soft grippers inspired by the human hand are designed to safely interact when grasping objects. This is achieved through the passive adaptability of the gripper and the flexibility of the soft material\cite{1}.
 They have some distinct advantages over hard and rigid robots. 
It is easy for them to adjust, pick and place delicate objects in randomized environments without damaging them, as well as the flexibility to operate in complex environments. 
In an attempt to make soft fingers more flexible and increase the range of options, researchers have tried many different actuation modes such as fluidic actuation\cite{2,3}, tendon-driven\cite{4,5}, shape memory metal\cite{6}, electronic stimulation\cite{7}, magnetic stimulation\cite{8}, and so on. 
These actuation modes produce different forces, response times, loads, and weights. However it is difficult to balance the output force, response time, and weight of soft grippers.

Researchers have proposed various methods of variable stiffness to address the weak load-carrying capacity with high response time in soft grippers. 
There are several directions for research: particle jamming\cite{9}, low melting point alloys\cite{10}, SMA\cite{11,12}, electric stimulation\cite{13}, additional exoskeleton\cite{14}, etc. 
These variable stiffness mechanisms have different variable stiffness ranges, response times, and scene layout requirements. 
They usually add rigid structures to the soft fingers. 
When the soft gripper comes into contact with a heavy object, the stiffness of the gripper increases thanks to the carried rigid structure activated by various controls.
These combinations preserve the compliance of the soft gripper and increase the load-bearing capacity. The addition of low melting point materials or SMAs into soft materials is one of the most straightforward methods, but longer solidification and melting times reduce their applicability\cite{15,16}.
Particle jamming or laminar jamming can also be effective in increasing the output force of the gripper, but the wide range of stiffness variations increases the size and weight of the gripper and also leads to unstable mobility of the particles\cite{17,18}.
The form of an additional exoskeleton can inherit the adaptability of soft materials and the high load-bearing capacity of rigid structures, such as the addition of ratchet and linkage structures. But it still makes the overall structure bulky\cite{19,20,21}.

This article introduces the Gripping Agent, which is specifically designed for grippers working in open-world environments. The agent integrates seamlessly with a Visual Language Model and complex light enhancement tools to improve cognitive abilities. A soft robotic gripper has also been investigated to achieve flexible grasping abilities, by evaluating its collision resistance and comparing finite-element analyses in mixed models. The key contributions of this article are as follows:
\begin{itemize}
    \item A soft gripper with variable rigidity has been designed to increase the versatility and gripping range of the gripper by varying its stiffness through motor actuation. This design innovation aims to boost the scope of objects the gripper can handle, and significantly improve its adaptability in various situations.
\end{itemize}
\begin{itemize}
    \item For the first time, a Visual Language Model has been integrated with a tool for brightness adjustments. The intelligent agent improves the soft gripper’s ability to grasp objects securely, even when in  complex outdoor environments. This innovative integration will enhance safety and reliability in soft robotic manipulations under unpredictable conditions. 
\end{itemize}
    \begin{itemize}
        \item The article promotes the use of an augmented monocular camera with VLM  to recognize object textures in order to increase the grasping success of soft robotic grippers. The object textures are divided into five categories, each of which can be recognized by the agent, effectively altering the gripper's rigidity through motor control. This further supports an adaptive way of lifting objects of different textures, improving the soft robotic gripper's overall load-bearing capacity.
    \end{itemize}
The remaining content of the article is below. Section III describes the design of the variable rigidity soft gripper, section IV analyzes the soft materials, springs, and tendons of the fingers, and section V evaluates the performance and stability of the variable rigidity gripper and demonstrates its capabilities experimentally. Finally, section VI summarizes the article.

\label{sec:headings}


\section{INTELLIGENT AGENT FOR VISUAL LANGUAGE MODELING OF SOFT GRIPPERS}

Lowlight Gripping Agent is composed of three basic components: The Prompt Engineer module; the Visual-Language Model core (VLM); and the Workflow Module. Figure 1 shows the framework for an advanced cognitive agent that incorporates a Visual-language Model equipped with an adaptive rigidity grasper, optimized for low-illumination. The innovation of the system lies in its integration of a large model with strong generalization capabilities and a low light enhancement algorithm \cite{zhang2023scrnet}. This allows it to cover a variety of scenarios while its soft-bodied grasper achieves precise and flexible object manipulation.

The Prompt Engine compiles task descriptions, role descriptions, tool descriptions and basic operational guidelines into a coherent prompt that can be processed by the VLM. It incorporates memory for optimizing prompt formulation to improve efficiency and effectiveness in image recognition and objects manipulation tasks, particularly in low-light situations.

The Visual-Language Model is an important part of the architecture. It converts complex instructions from Prompt Engineers into directives the machine can understand within Workflow.

The Workflow module uses the advanced functionality of GPT-4.0 and applies the Chain of Thought method to enhance the perception of objects. This allows for more accurate object detection and the determination of the best points for grasping. Agents initially use advanced image processing to improve visibility and clarity in images acquired under dimly lit. The second step is to identify the objects that need to be manipulated. It is important to distinguish between the object intended and the grasping device, and also consider the size, location, and material composition of the target object. This critical stage is required for the system's grasping to work properly. After detecting the object the system must calculate the best to grasp based on different physical characteristics. The complex spatial analysis includes determining the center of mass, the surface geometry, and the best angle to approach the gripper. The robotic gripper is used to physically grab the object in the fourth step. The gripper differentiates between successful operations (labeled "Correct Results") and errors (labeled "Abnormal Error"). The historical data is then used to enable adaptive learning. This allows the system's algorithms to be improved and its object-handling abilities to be enhanced over time.

\begin{figure}
  \centering
  \includegraphics[width=0.8\textwidth]{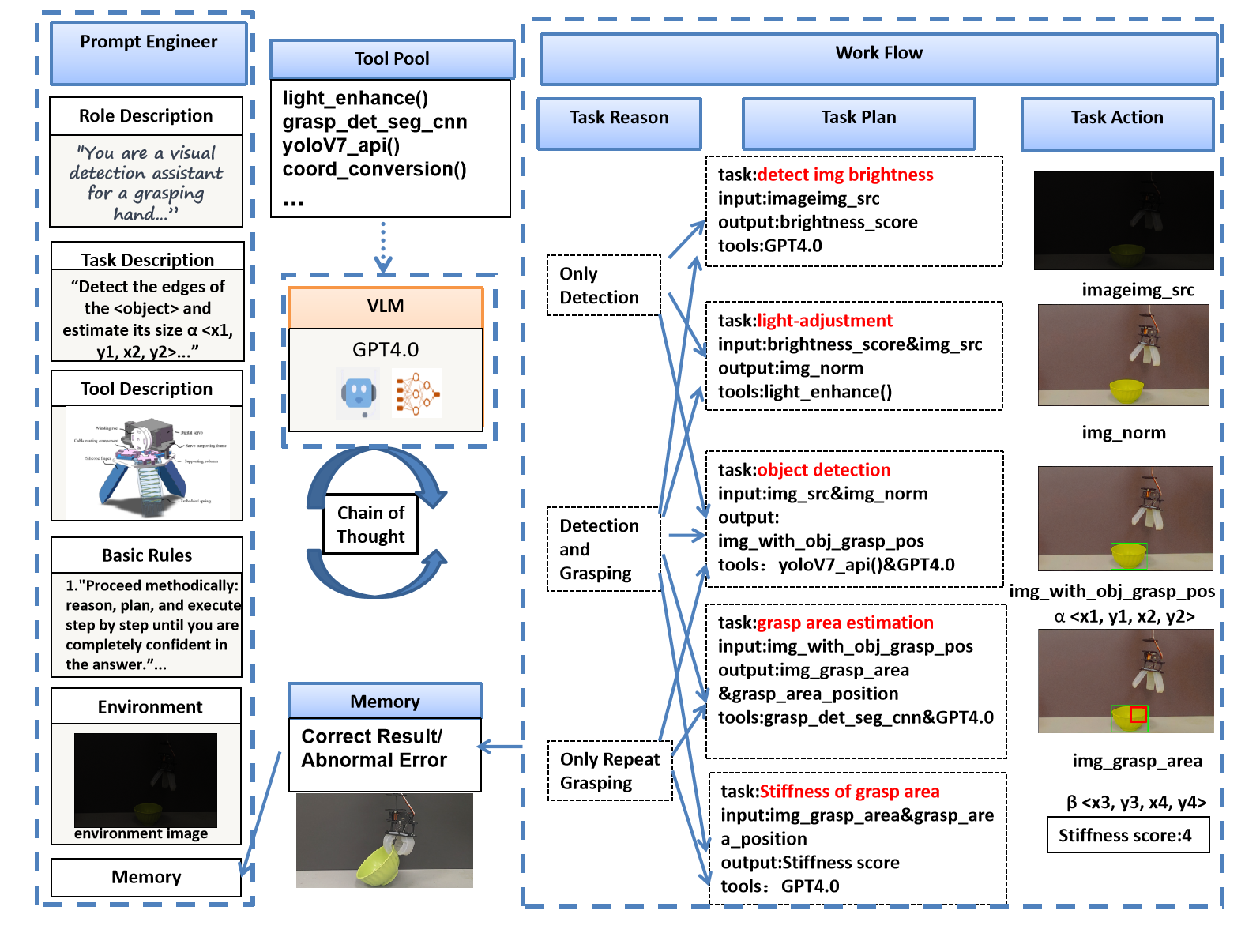}
  \caption{LGAgent (Lowlight Gripping Agent) consists of three key components: the Prompt Engineer, the Visual-Language Model (VLM), and the Work Flow.}
  \label{fig:LGAgent}
\end{figure}

In summary, this intelligent agent system processes images to enhance physical details such as materials, contours, and positions. In conjunction with the Thought of Chain Strategy, the VLM distinguishes the object being grasped and the gripper.  Furthermore, the stiffness of the soft gripper, as well as the grasp area, is individually adjusted based on the materials and contours of the object to ensure successful gripping. 
The system continuously improves its grasping technique by recording outcomes throughout this iterative procedure,  which enhances the gripper's adaptability and ability to generalize. Finally, LGAgent’s end-to-end learning approach also improves the gripper's ability to handle grasping tasks that involve irregular objects, different materials, and novel scenarios.

\section{Design of bionic hybrid gripper}

The design concept of the bionic hybrid gripper is derived from the human hand. The human hand has evolved over the years to become incredibly flexible, whether it's grabbing small items or carrying large ones. When different shapes of objects need to be grasped, people use different ways of grasping, such as pinching, enveloping, hooking, lifting and other actions. When it comes to grasping objects that have never been seen before, the human hand also does a good job of holding onto things without damaging them because of the natural structure of the human hand, as shown in Figure 1. The human hand is composed of tendons, hand bones and flesh. The human hand bone is the frame and is mainly responsible for supporting and connecting various parts and providing bearing capacity. The main function of the meat is cushioning and adaptive ability, when the hand touches the object, the meat can to some extent unload part of the force and increase the contact friction according to the surface of the object. The human hand is powered by the expansion and contraction of tendons. In addition to driving the hand, the tendon allows the hand to bend at an angle in the opposite direction, which also allows the hand to grasp larger objects. 

\begin{figure}
  \centering
  \includegraphics[width=0.6\textwidth]{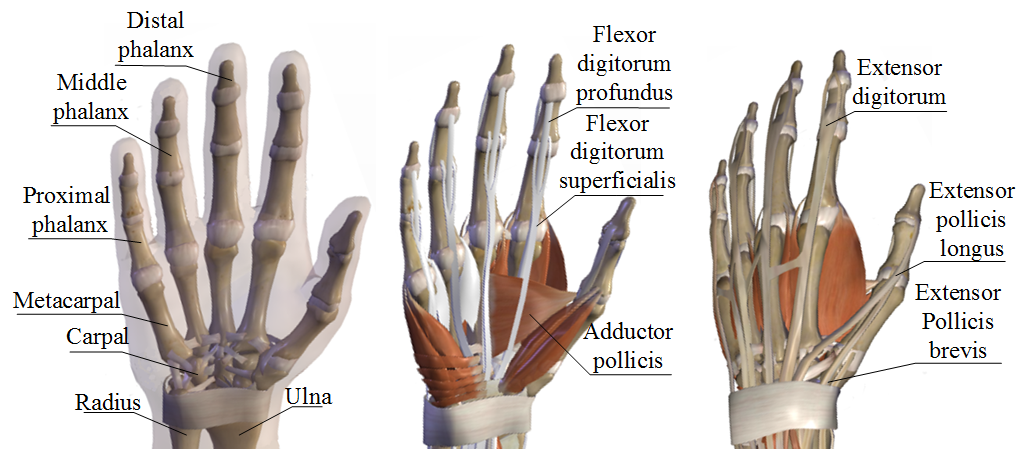}
  \caption{Anatomy of the human hand}
  \label{fig:anatomy_of_hand}
\end{figure}

On the basis of the research on the hand, the authors propose a new rigid-soft combination of soft gripper, which is composed of rectangular springs, silicone, cables and a 3D printed frame, as shown in Figure 3. Rectangular springs are designed to mimic the functions of the hand bone, providing stiffness without compromising the softness of the soft body when gripping an object. In addition, The spring can be stretched by a separate motor to increase the stiffness of the spring, thus increasing the total load capacity of the hybrid gripper. Silicone mimics the flesh of the human hand and has the ability to passively adapt when interacting with objects. It is worth noting that a single cable is unstable as a medium in the case of a motor driving three fingers. So the number of cables is doubled so the fingers will move according to our settings when bending. The cable is mounted is both free ends are mounted on the rollers of the gripper and pass through the rectangular spring and the hole in the end of the fingertip.
\begin{figure}
  \centering
  \includegraphics[width=0.6\textwidth]{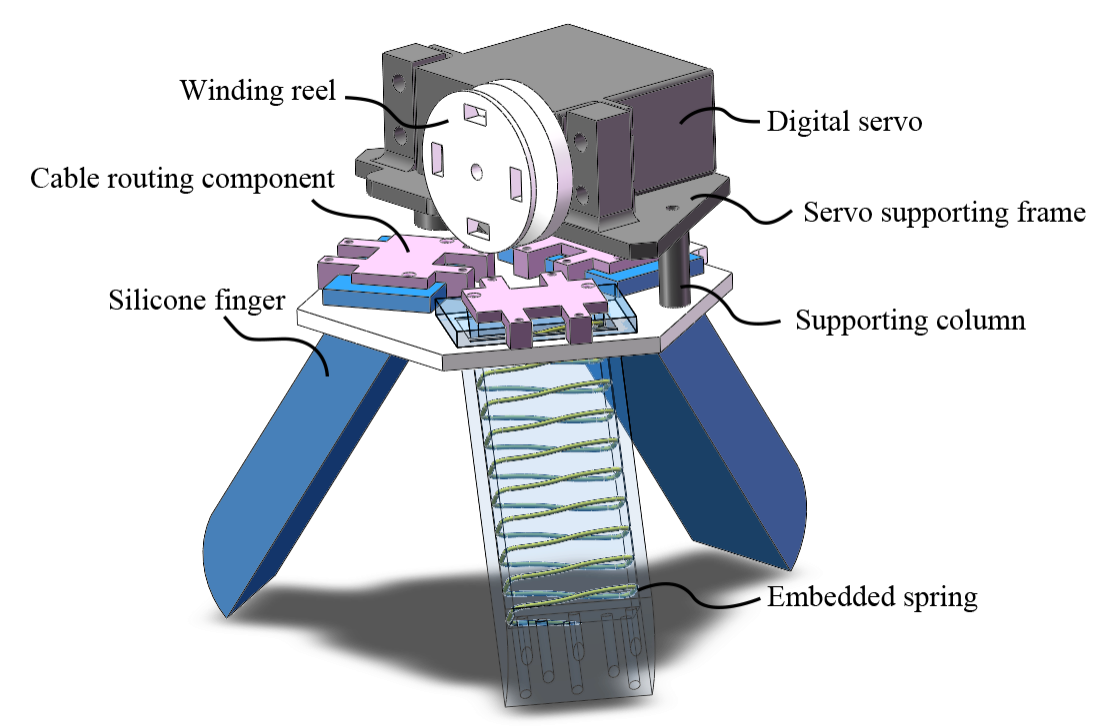}
  \caption{Three-finger hybrid gripper construction}
  \label{fig:three_finger_gripper}
\end{figure}
The spring embedment isolates the contact between the cable and the silicone into friction between the cable and the spring. The embedding of the spring isolates the friction between the cable and the silicone and becomes the friction between the cable and the spring. This greatly reduces the amount of energy lost by the tendon during movement, thus increasing the output force of the finger.

\begin{figure}
  \centering
  \includegraphics[width=0.8\textwidth]{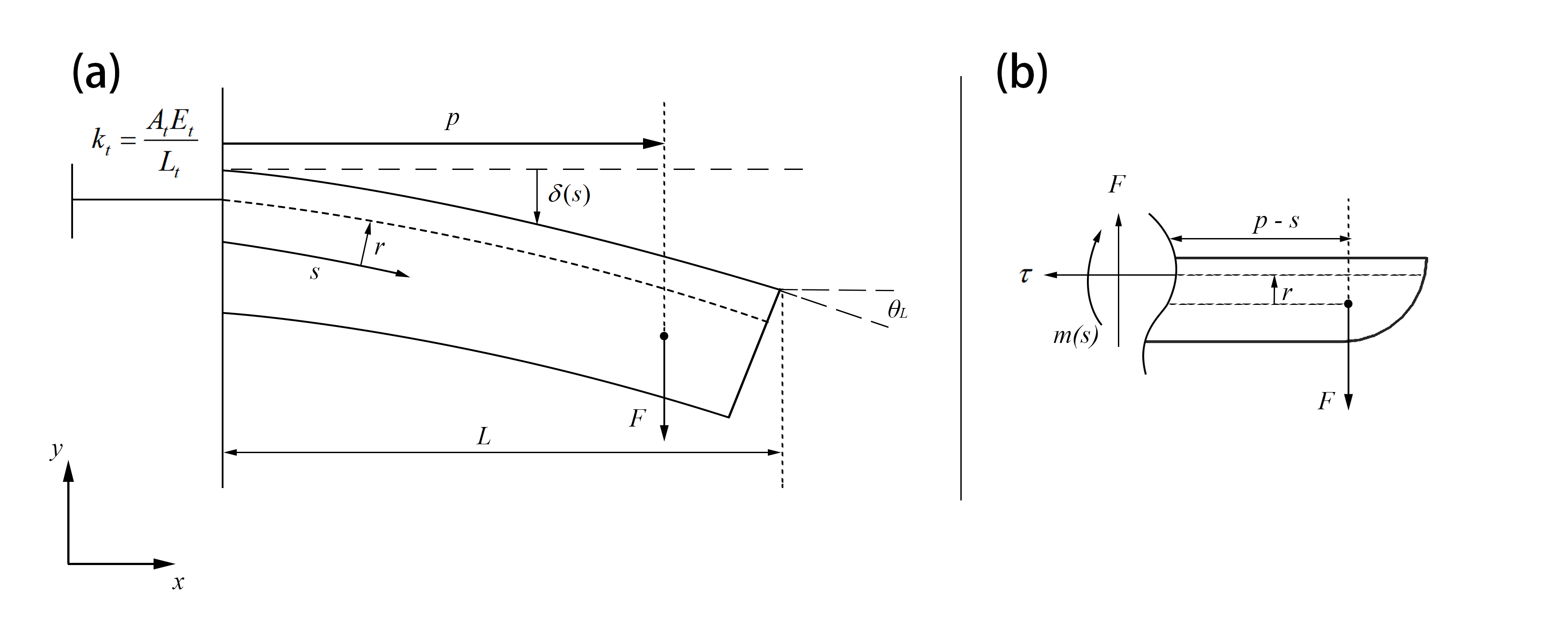}
  \caption{Simplified schematic. (a) Simplified diagram of the gripper as a whole, (b) forces on the fingertips. }
  \label{fig:deformation_diagram}
\end{figure}

\section{ANALYSIS OF TENDON PATH AND SPRING ON FINGER STIFFNESS}
The advantage of the tendon drive is that the actuator and robot structure can be separated, allowing for a more compact and flexible robot design. 
In the following we analyze the effect of parallel and converging tendons on the stiffness of the gripper. 
The author use classical beam theory to construct mathematical models of tendons and hybrid finger. 
Since all three fingers have the same design, we will only analyze one finger here. 
First, since it is mentioned above that the friction between the cable and the spring is minimal so the friction between them is ignored. 
The fingers are simplified as shown in Figure 4.

Based on Fig. 3, the static moment equation can be written out, and and by variation an expression for the internal moment m(s) and the arc length, and according to the linear constitutive law, we obtain

\begin{equation}
    \begin{aligned}
        \begin{cases}
             m(s) = 
             \begin{cases}
                F(s-p)+r\pi & \ s<p \\            
                r\tau       & \ s \geq p \\
             \end{cases}\\
         m(s) = - EI \frac{d\theta}{ds}   
        \end{cases}         
    \end{aligned}
\end{equation}

Integrating the equation with the boundary condition \( \theta(0)=0 \)and enforcing continuity in \( \theta(s) \) at s = p, we can get
\begin{equation}
    -EI\theta(s) =
    \begin{cases}
        \frac{1}{2}Fs^2+(r\tau-Fp)s & s<p \\
        r\tau s -\frac{1}{2}Fp^2 & s \geq p \\
    \end{cases}
\end{equation}

The tendon reaction tension \( \tau \) can be obtained from the tendon stretch \( \Delta\) due to finger deflection and the spring constant of the tendon.  
 \begin{equation}
     \Delta = r \theta _L \\
 \end{equation}
 \begin{equation}
     \tau = k_t r \theta_L\\
 \end{equation}

When s = L and combining equations (2) and (4), we can obtain
\begin{equation}
    \begin{cases}
        \theta_L = \frac{Fp^2}{2(EI+r^2k_tL)} \\
        \tau = \frac{Fk_trp^2}{2(EI+r^2k_tL)} \\
    \end{cases}
\end{equation}

Approximating the small angle as \( \theta \approx \frac{d\delta}{ds} \), by integrating again and reinforcing the boundary \(\Theta(s)\) at s = p to get
\begin{equation}
    -EI\delta(s)
    \begin{cases}
        \frac{1}{6}Fs^3+\frac{1}{2}(r\tau - Fp)s^2 & s < p \\
        \frac{1}{2}r\tau s^2-\frac{1}{2}Fp^2s+\frac{1}{6}Fp^3 & s \geq p
    \end{cases}
\end{equation}

Where \(\delta\) is deflection. When the tendon is immobilized at the fingertip (p = L), the deflection of the fingertip can be written as
\begin{equation}
    \delta_{tip} = \frac{Fp^3}{3EI}-(\frac{r^2k_tL}{EI+r^2k_tL})\frac{FL^3}{4EI}
\end{equation}

From Equation (10), it can be seen that when there is no tendon(k = 0) or the tendon is mounted in the middle layer(r = 0), the deflection of the fingertip is the general beam deflection formula. 
Meanwhile, we can know that it is possible to increase r and k to increase the stiffness of the fingertip, and that increasing r is more effective than increasing k.

For tendons with sufficiently large stiffness we can
approximate $  \lim\limits_{k_t\rightarrow \infty} \theta_L = 0 $, then we get
\begin{equation}
    \delta(p) = \frac{Fp^3}{3EI}-(\frac{p}{L})\frac{Fp^3}{4EI}
\end{equation}

This equation represents the deflection of the parallel tendon, and by adding the slope a and distance between the end of the tendon and the root b to the parallel tendon.
we can obtain
\begin{equation}
    -EI\theta(s) =
    \begin{cases}
        \frac{1}{2}(F+a\tau)s^2 + (b\tau - Fp)s & s<z \\
        \frac{1}{2}a\tau s^2 + b \tau s - \frac{1}{2}Fp^2 & s \geq z
    \end{cases}
\end{equation}

In the same way as for the calculation of the parallel tendon, and when the fixed point at the end of the cable is at the end of the finger (p = L), we get
\begin{equation}
    \delta_{tip} = \frac{FL^3}{3EI}-\frac{\tau}{6EI} (aL^3 + 3bL^2)
\end{equation}

The total tendon stretch can be written as
\begin{equation}
    \triangle = \int_{0}^{L} r(S)(\frac{d\theta}{ds})ds
\end{equation}
where \( \frac{d\theta}{ds} = -\frac{m(s)}{EL}\)  and \( \tau = k \triangle\), and the tendon reaction tension can be obtained
\begin{equation}
    \tau = \frac{k_tF(aL^3+3bL^2)}{6EI+k_tL(2a^2L^2+6abL+6b^2)}
\end{equation}

Combining Equations 10 and 12, the tendon reaction tension on convergent tendon can be written as
\begin{equation}
    \delta_{tip}= \frac{FL^3}{3EI}-(\frac{{aL+3b}^2}{3(a^2L^2+3abL+3b^2)})\frac{FL^3}{4EI}
\end{equation}

The end deflection angle of which can be expressed as:
\begin{equation}
    \theta_L = \frac{FL^2}{2(EL+r^2Lk_t)}
\end{equation}
where \( k_T\) can be written as:
\begin{equation}
    k_T = \frac{k_s k_t k_m}{k_t k+m + k_s k_m + k_s k_t}
\end{equation}
\( k_T\) is the total stiffness coefficient, \( k_s\) is the stiffness coefficient of the spring, kt is the stiffness coefficient of the tendon and \( k_m\) is the stiffness coefficient of the soft material.
Comparison of the effects of pure silicone, silicone with parallel tendon and silicone with converging tendon on finger stiffness is shown in Figure 5.

\section{EXPERIMENTS}

\subsection{ Variable stiffness analysis of finger }
First, we dimensionlessly normalized the formula to compare pure silicone, silicone with parallel tendon, and silicone with convergent tendon. 
Maximizing the slope of the convergent tendon to -1 gives us Figure 5. 
From the figure we can clearly see that with the addition of the tendon, the tip deflection will be less than in a pure silicone finger, while the convergent tendon is stiffer than the parallel tendon.

\begin{figure}
  \centering
  \includegraphics[width=0.8\textwidth]{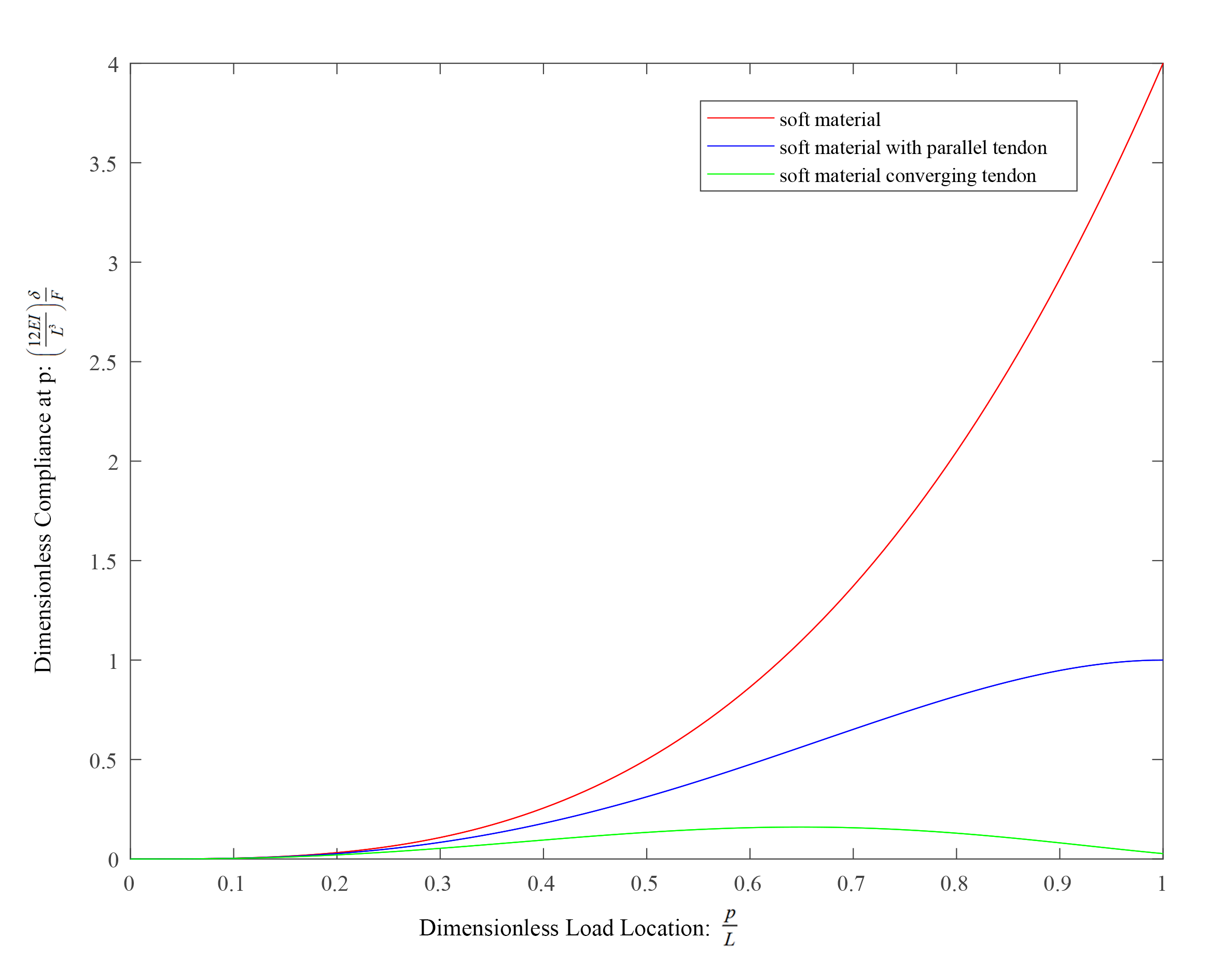}
  \caption{Comparison of fingertip stiffness of pure soft material, soft material with parallel tendon, and soft material with convergent tendon. The horizontal coordinate represents the end position of the tendon mount and the vertical coordinate represents the deflection of the end of the finger. The red, blue, and green lines represent fingertip deflection for pure soft material, soft material and parallel tendon, and soft material and convergent tendon, respectively.}
  \label{fig:deformation_diagram}
\end{figure}

\begin{figure}
    \centering
    \subfigure(a){\includegraphics[width=0.3\textwidth]{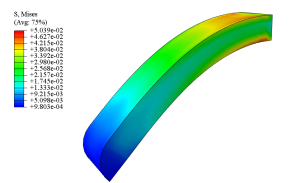}}
    \hfill
    \subfigure(b){\includegraphics[width=0.3\textwidth]{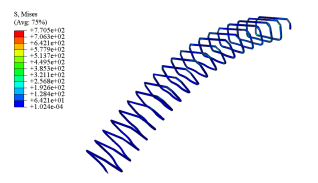}}
    \hfill
    \subfigure(c){\includegraphics[width=0.3\textwidth]{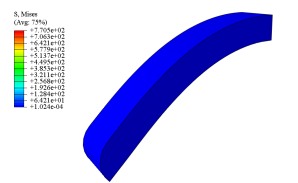}}
    \caption{Stresses on the three types of structures}
    \label{fig:horizontal_layout}
\end{figure}

\begin{figure}
  \centering
  \includegraphics[width=0.8\textwidth]{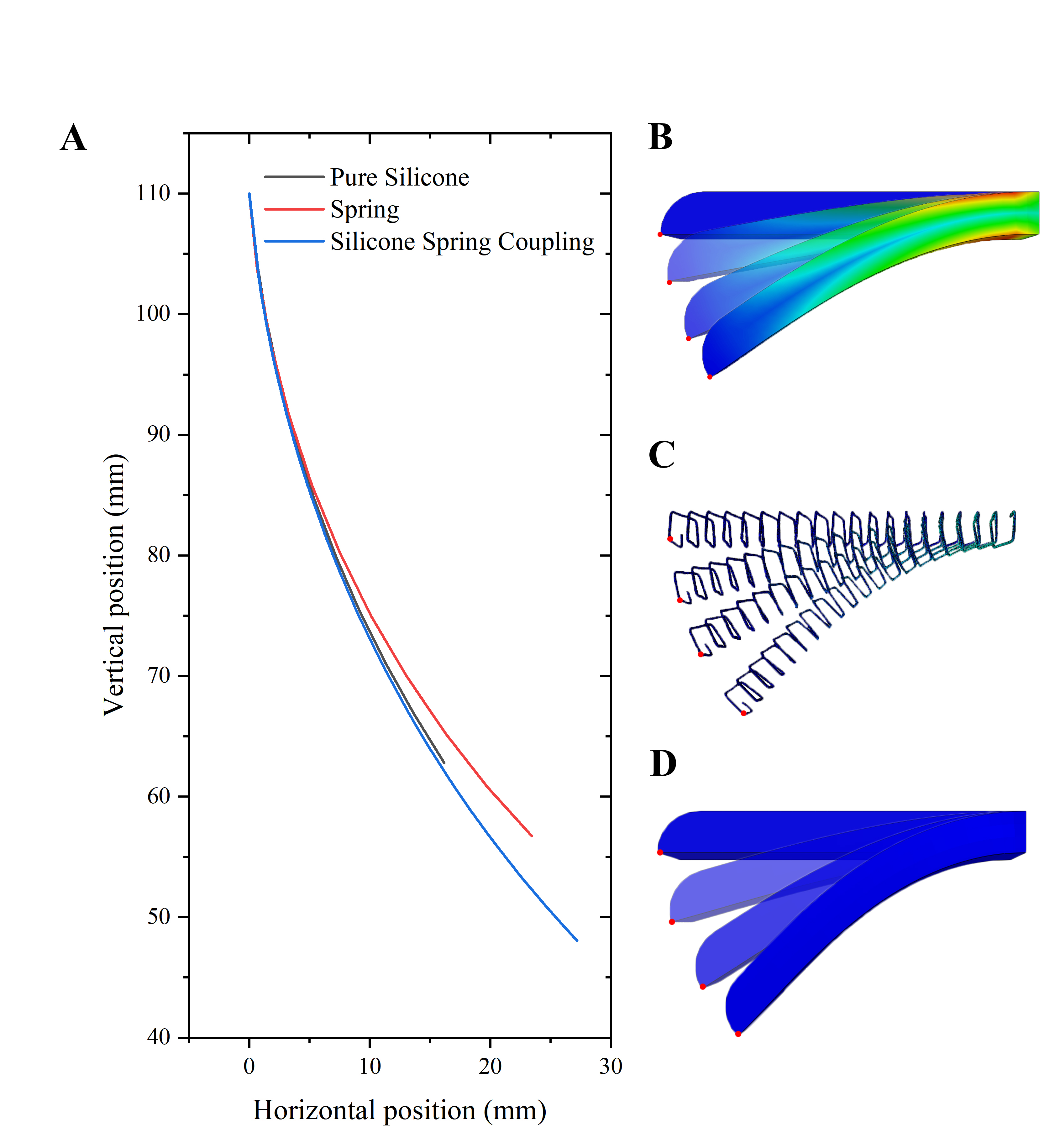}
  \caption{Displacement motion of coupled fingers. (A) Displacement curves for pure silicone, spring and spring-silicone coupled finger. (B) The movement process of pure silicone fingertips. (C) The process of movement of the end of the spring. (D) Silicone spring coupled finger movement process.}
  \label{fig:deformation_diagram}
\end{figure}

\begin{figure}
  \centering
  \includegraphics[width=0.8\textwidth]{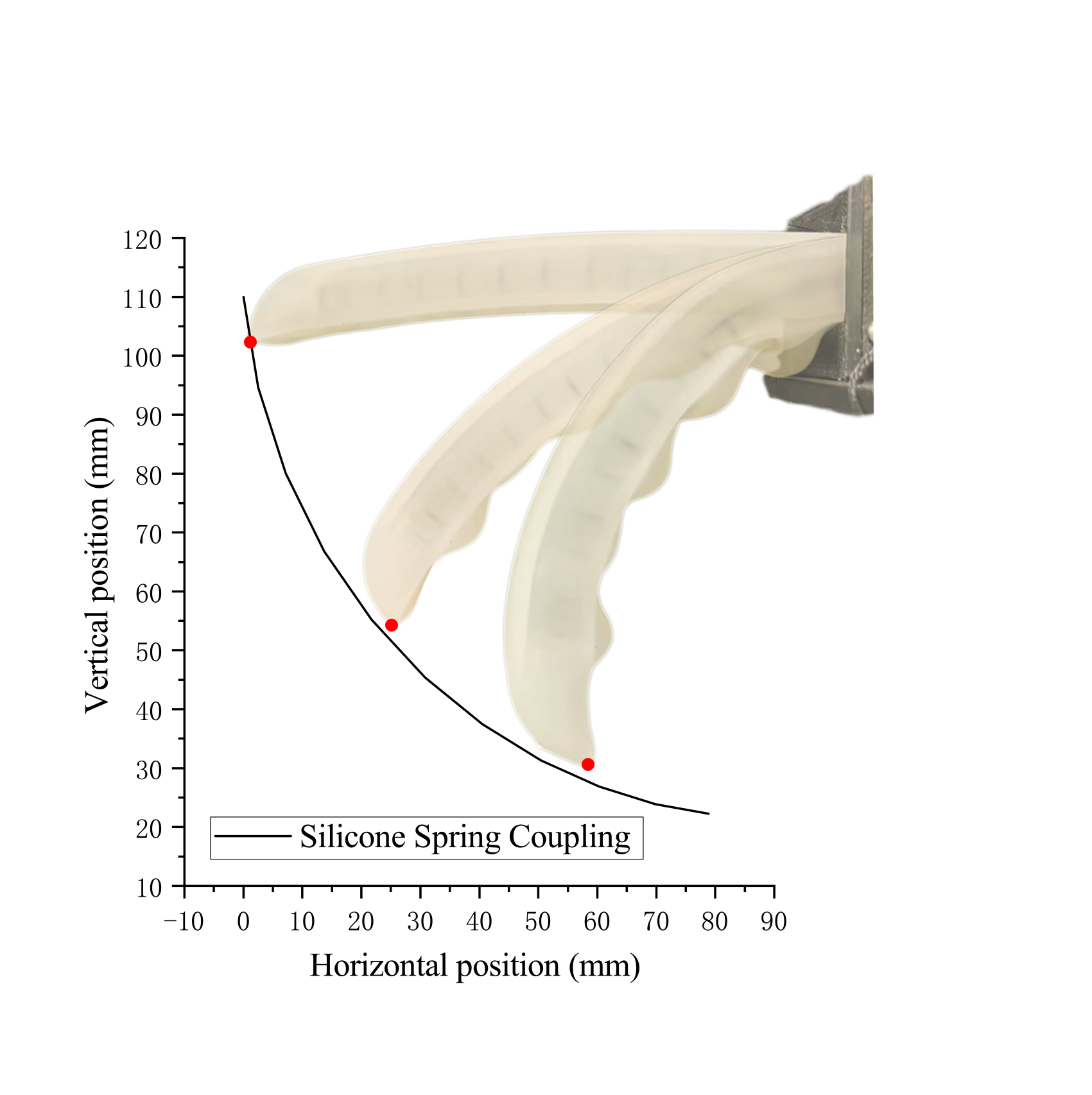}
  \caption{Displacement of the simulated coupled finger compared to the displacement of the physical finger. The black line is the simulated trajectory and the red dots are the 0°, 45° and 90° coordinates of the motor movement.}
  \label{fig:deformation_diagram}
\end{figure}
In grasping experiments with all three structures, we find that the convergent tendon can pick up equally heavy objects more easily and that it has a higher load carrying capacity. This is due to the mounting position of the tendon and the fact that the converging tendon can complete the grasp with little to no contact with the spring.
However, we set the slope of the convergent tendon to -1 (maximum) which means that when the tendon is
mounted in the middle layer, the finger bends in the opposite direction under an external force. In other words,when the convergent tendon is at its maximum value, the gripper can only grasp objects smaller than the mounting range, and when pinching is used to grasp a large object, the external force generated by the object on the gripper will cause the gripper to bend in the opposite direction under tendon actuation and the grasp will fail. 
Thus converging the tendon to take the maximum value is theoretically optimal, but not for the design of the gripper, whereas choosing the mounting position of the tendon to be infinitely close to the middle layer without reaching the middle layer does not cause the problem.

Simultaneous finite element analysis of pure silicone, springs, and hybrid finger with silicone and spring, we can see that under the same tensile force, a hybrid finger embedded in a spring structure allows the stress on the silicone to be dispersed evenly over all areas. 
This not only provides a controlled grasp, but also allows for safer interaction with objects.
By comparing the displacement curves of the structures, the trajectory of the hybrid finger is more linearized, as shown in Figure 7. 
The pure silicone finger stops after moving vertically to a position of 45mm. This may be due to the fact that Dragon Skin 10 is incompressible, resulting in a large reverse force on the substructure of the finger in this position, which causes the finger not to be able to bend. 
The spring silicone coupling finger has a maximum vertical displacement (62 mm) while the spring bends up to 55 mm. Comparing the vertical and horizontal displacements of the three, the bending curvature of the spring-silicone coupled finger is more linearized.

To verify the accuracy of the simulation, the simulated tension is increased and compared with the actual bending. 
At the same time, we record the three coordinate positions of the actual finger trajectory corresponding to $0^\circ$, $45^\circ$, and $90^\circ$ of motor rotation to compare with the simulation curve. 
The resulting displacement comparison plot, shown in Figure 8.

As we can see from the figure, the reason why the initial position is not at (0, 110) is because the tendon has a preload on the finger during mounting. There is a preparation of the finger toward bending under the preload, so the finger will bend a certain distance.
Secondly, since the finger is subjected to gravity during the experiment, this also results in more horizontal displacement of the actual finger than that of the simulation. 
Thirdly, when bending, the lower surface of the actual finger forms a crease due to the incompressibility of the silicone, which leads to the discrepancy between the actual displacement and the simulated displacement. 
Overall, the comparison of the bending results shows a more regular bending trajectory of the finger.

\subsection{ Stabilization of variable stiffness finger }
\begin{figure}
	\centering
	\includegraphics[width=\columnwidth]
	{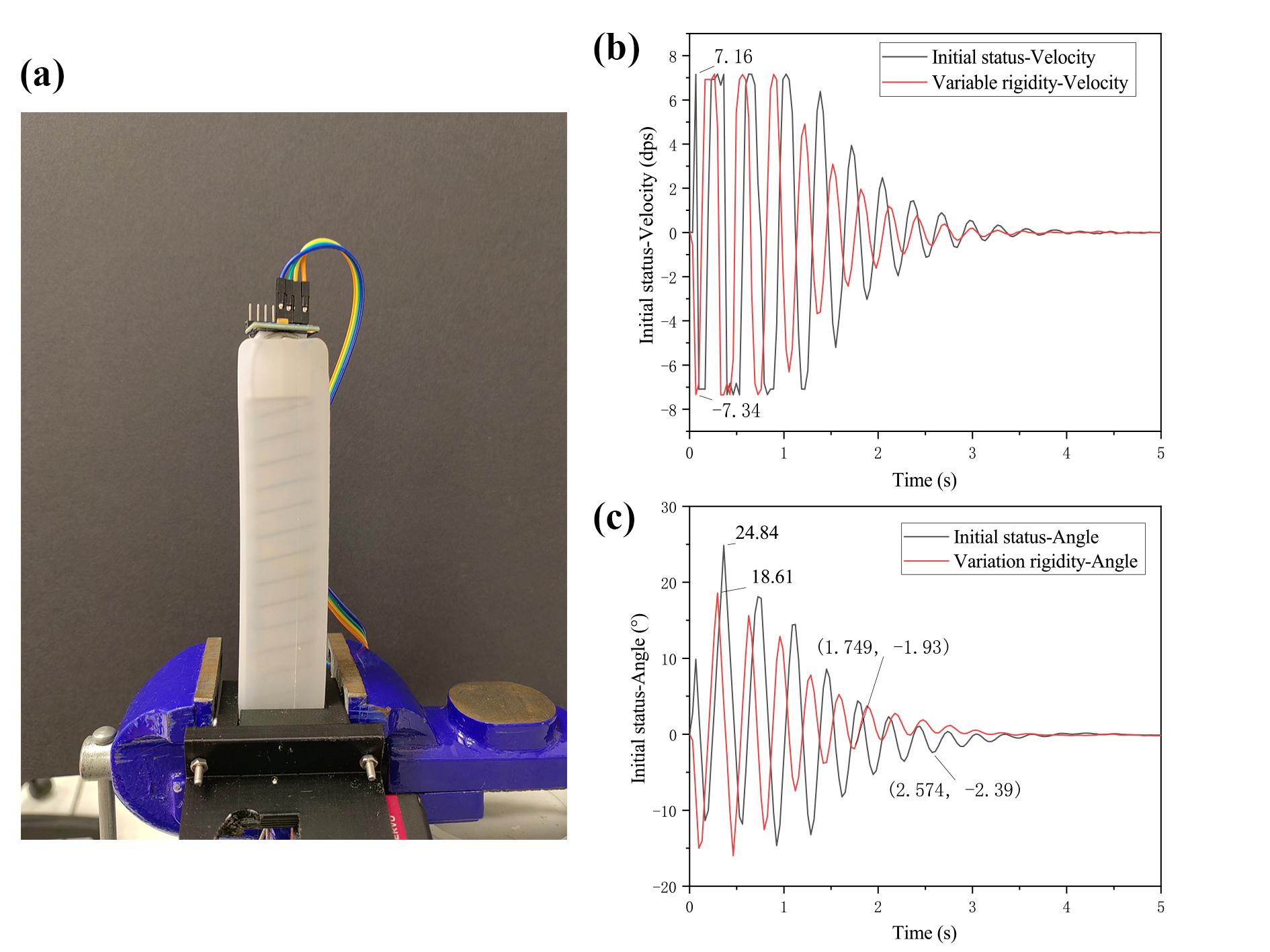}
	\caption{Rigid-soft finger stability test. (a) The finger is mounted in a bench vise and the IMU at the fingertip collects the angle and velocity signals from the finger after it is struck. (b) Velocity change graph for the initial state and after pretensioning. (c) graphs of velocity changes in the initial state and after pretensioning.}
	\label{fig:control}
\end{figure}
The stability of a gripper is an equally important indicator of a gripper's performance. 
When grippers are used in industry or on robots, it is inevitable that they will encounter collisions that will affect the equipment. 
When grippers are used in drones, for example, the fingers can vibrate due to air resistance or external disturbances. 
Such vibrations can be hugely disruptive to the control of a drone, so it is a prerequisite to be able to stabilize it quickly after a similar situation is able to be widely used.
To expand the applicability of rigid-soft grippers in a variety of fields, the authors conducted interference stability experiments on the grippers, comparing the time to regain stability before and after spring pretensioning,and the angular range of change.

In this experiment, we mounted the MPU6050 at the end of the fingertip and rigid-soft finger is mounted in a bench vise, as shown in Figure 9(a). We collected data from IMU by repeatedly tapping finger for comparison.

Curves with similar initial angular velocities are selected for comparison to compare the accuracy of the experiments. 
Two experiments with an initial state angular velocity of 7.16 dps and a starting angular velocity of 7.34 dps after pretensioning are selected. 
We can clearly see that the pre-tensioned finger returns to its steady state more quickly despite the fact that it was loaded with a higher initial velocity. 
From Figure 9(c), it is clearly seen that the amplitude of the finger that has not been pretensioned is much larger, with a maximum angle of $25^\circ$ on one side, whereas the finger that has been pretensioned is only about $19^\circ$. 
Here we consider an amplitude of less than $3^\circ$ as reaching a steady state. 
The finger that was pretensioned reaches a stable state at 1.8 s, while the finger in its original state takes 2.5 s to reach a stable state. 
Consequently, rigid-soft fingers return to a steady position more quickly after tightening, which makes it easier to design control systems and perform faster operations for applications such as mobile robotics, industrial production, and domestic application.

\subsection{ Grasping validation }

\begin{figure}
  \centering
  \includegraphics[width=0.8\textwidth]{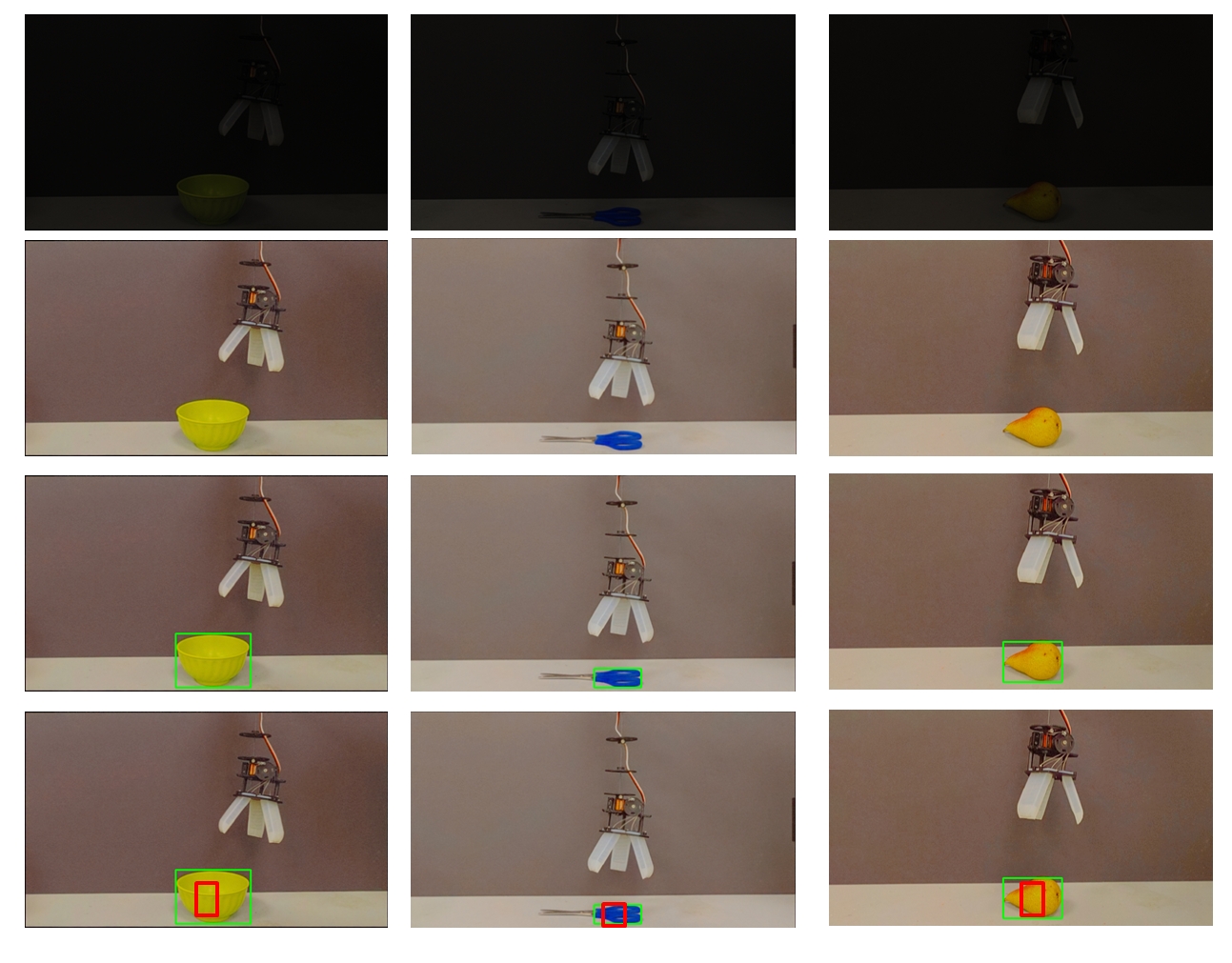}
  \caption{Demonstration of Workflow section. The top row exhibits images captured under low-light conditions. The second row presents the same scenes with enhanced brightness by using a low-light enhancement algorithm. The third row marks the identification and localization of target objects for grasping. Finally, the fourth row illustrates the estimation of the optimal gripping area, denoted by a red box.}
  \label{fig:Workflow}
\end{figure}

\begin{figure}
  \centering
  \includegraphics[width=0.6\textwidth]{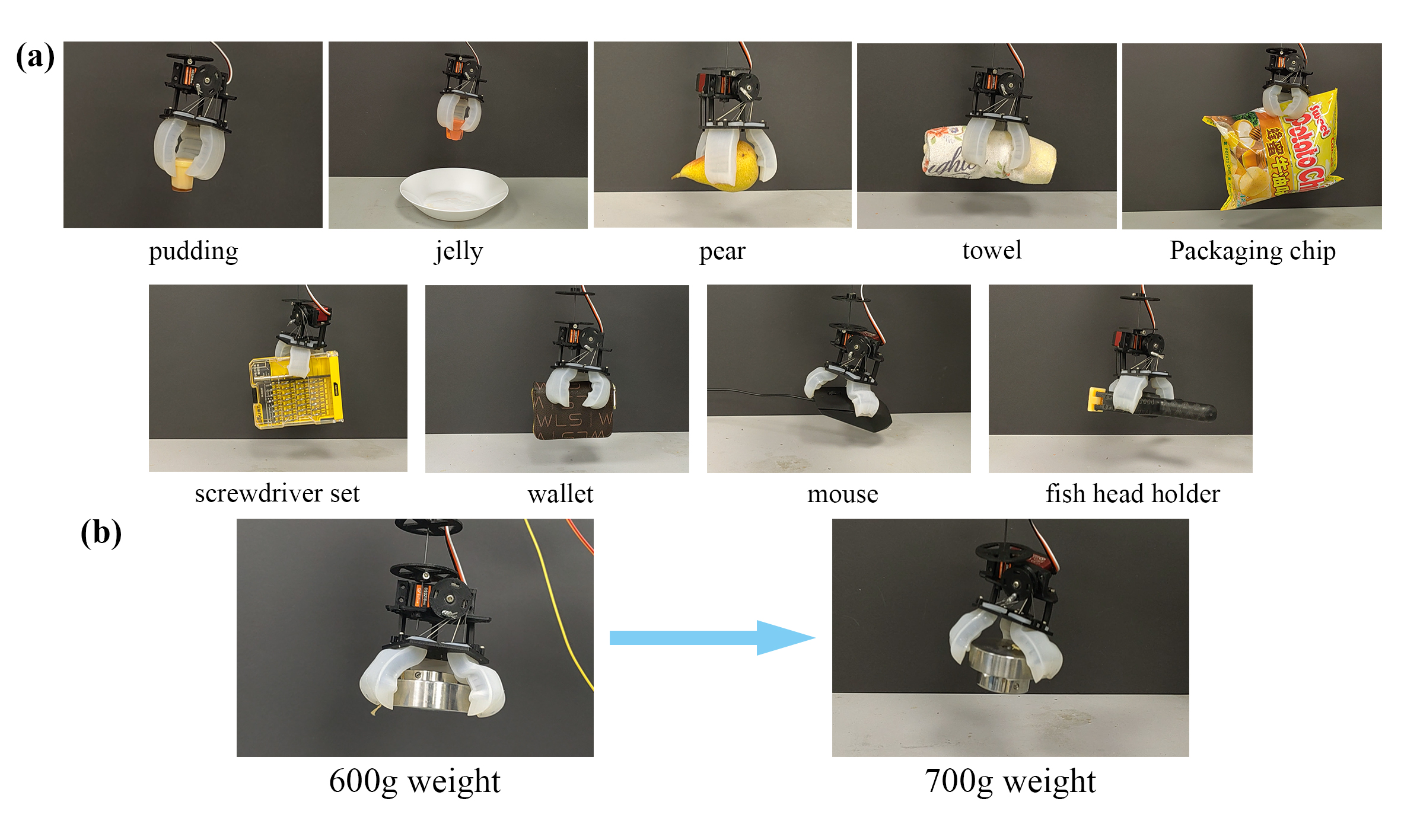}
  \caption{(a) Items for training grasping (b) The embedded springs are pre-stretched to enhance the load-carrying properties of the gripper.}
  \label{fig:three_finger_gripper}
\end{figure}

\begin{figure}
  \centering
  \includegraphics[width=0.8\textwidth]{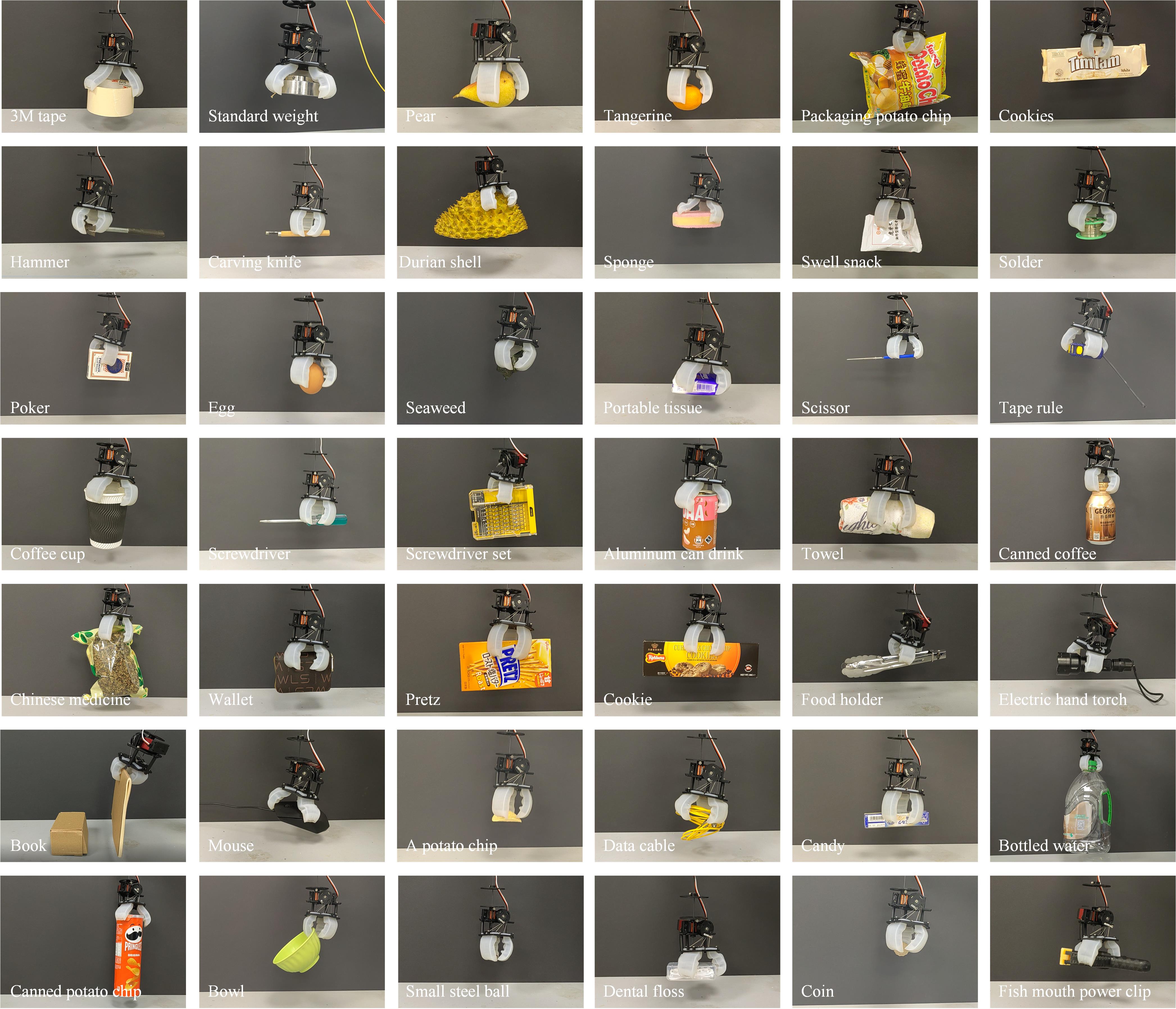}
  \caption{42 daily necessities and industrial tools.}
  \label{fig:three_finger_gripper}
\end{figure}

The gripper is designed as a three-fingered circular array, to accommodate the gripping pattern of the drone in the grab. 
The gripper is mounted on a section of a continuum robotic arm for grasping experiments and random time periods are to simulate complex scenarios under different offsets outdoors.
The goal is to mimic a variety of complicated real-world scenarios with unpredictably shifting conditions outdoors.
During these experiments, our innovative gripper automatically can adjust its rigidity. By integrating object recognition feedback from the single-lens camera with the VLM model, it enables appropriately responding to diverse grasping dynamics.

We divide the rigidity of the gripper from soft to hard corresponding to 0-4 into 5 classes, with 0 being the softest original state and 4 being the hardest heavy grasping state. The hardness level as well as the material of the captured item is categorized as shown in Table 1.

\begin{table}[!tb]
	\centering
	\caption{Hardness classification of soft gripper.}

\begin{tabular}{|p{2.3cm}|p{2.3cm}|p{2.3cm}|p{2.3cm}|p{2.3cm}|p{2.3cm}|}
\hline
Stiffness Levels &
  0 &
  1 &
  2 &
  3 &
  4 \\ \hline
Objects &
  Vulnerable items such as jelly, potato chips, persimmons, etc. &
  easily deformed items such as fruits, plastic packaging, etc. &
  Tough and easily deformed items such as leather, towels, etc. &
  Hard plastic or wooden items &
  Heavy objects such as weights, dumbbells, and other metal objects \\ \hline
\end{tabular}
\label{tab:Hardness}
\end{table}

A miniature motor has been integrated to variably adjust the gripper's rigidity, along with a refined VLM model embedded with an automatic luminosity modification tool. Whenever a gripping task is initiated, if the available natural light is excessively bright or overly dim, the VLM model activates the brightness augmentation tool. This aids in enhancing an item's outline and material recognition, which supports the selection of a suitable gripping style and stiffness adjustment. The modification process is illustrated in Figure 10.
The VLM model provides real-time brightness modification proportionate to the intricate environmental conditions. Subsequently, it recognizes potential items for grip and determines the best gripping zone within its visual range.

In the preliminary stage, ten items with varying features were chosen to train the VLM model, including those with regular and irregular shapes and hardness levels ranging from 0 to 4. These selected items in increasing order of hardness are - jelly, pudding, packet of chips, pear, towel, leather wallet, screwdriver set, mouse, fish tail clip, and weight, as illustrated in Figure 11(a).
We conducted 20 rounds of gripping tests on these items over three distinctive timeframes - normal light conditions in the morning and afternoon, overexposure at noon, and low light conditions in the evening. During the training process, the observations revealed that the intelligent system without employing the brightness adjustment tool failed to focus properly on any given object for gripping during noon and evening, which led to a failure in gripping.
In dealing with smooth surfaces like leather and hard plastics, the gripper, at hardness level 0, upholds a steady grip with a 90

In our pursuit to evaluate the robustness of the established VLM model, we conducted extensive experiments involving 45 diverse objects\cite{22}. These objects were chosen to mirror the breadth of real-world scenarios, and consequently, offered a comprehensive platform to assess the competency of our adjustable gripper.
Our gripper demonstrated an intelligent adaptability, marking a successful grasp of all tested objects without inflicting any damage. This benchmark performance can be attributed to the dynamically versatile rigidity adjustments of the gripper, facilitated by the VLM model. The model, in real-time, processed visual feedback to determine the optimal gripping style.
To rule out randomness and lend more credence to our experiment, each grasping task was performed twice, at different intervals. The trained VLM model was effective and precise for most items, ensuring a smooth grasp. A noteworthy incident was when the VLM model initially misclassified a durian shell as a fruit, setting its hardness level at 1 and subsequently failing the task. However, the responsive system quickly rectified the error, appropriately reassigning the hardness level at 3, thereby achieving a successful grasp of the durian shell.
Consolidated results of these experiments are depicted in Figure 12. They clearly emphasize the capability of our adaptable gripper in successfully performing complex grasping tasks in dynamic outdoor environments when fused with a robust visual language model. This testifies the grasper's potential for real-world applications, reflecting promising prospects for the gripper's replicability and scalability.

\section{Conclusion}
In this research, GAgent is a VLM-based soft gripper with variable rigidity designed for open-world situations. A monocular camera combined with VLM and light-enhanced tools allows for the recognition of texture, grip area, and location of objects, improving the success of soft robotic graspers. In addition, this study proposes a bionic hybrid gripper that utilizes a combination of variable rigidity structures and soft materials to provide the gripper with excellent flexibility and load carrying capacity. 
Finger stiffness with tendon, spring and soft materials is investigated and discussed and finite element analysis is performed.
Through a series of experiments, it has been proved that the lightweight design of the gripper has outstanding loadbearing capacity and softness. 
Not only does it interact
safely with everyday objects, it also interacts well with
thorny objects and extremely soft objects. 
This gripper shows its potential in the field of drones, such as transportation and detection in extreme conditions. 
In the future, more flight application experiments will be conducted and its performance will be further enhanced through material improvements and increased sensing capabilities.

\section*{Acknowledgments}
The work described in this paper is partially supported by Shenzhen Fundamental Research (General Program)(WDZC20231129163533001) 

\bibliographystyle{unsrt}  
\bibliography{references}

\end{document}